\documentclass{article}

\usepackage{arxiv}

\usepackage[utf8]{inputenc}
\usepackage[T1]{fontenc}
\usepackage{hyperref}
\usepackage{url}
\usepackage{booktabs}
\usepackage{amsfonts}
\usepackage{nicefrac}
\usepackage{microtype}
\usepackage{graphicx}
\usepackage{natbib}
\usepackage{doi}
\usepackage{tikz}
\usetikzlibrary{positioning}

\title{StocksTalk: A Voice-Enabled Conversational Agent for Structured Query Generation over Web Data}

\renewcommand{\shorttitle}{StocksTalk}

\newif\ifuniqueAffiliation
\uniqueAffiliationtrue

\ifuniqueAffiliation
\author{
Akshat Parmar\thanks{Equal contribution. \\  
\href{https://github.com/Finance-LLMs/StocksTalk}{Code} \quad | \quad
\href{https://drive.google.com/file/d/10QtXvr-usbmAjzC5bbd54Lu51_KaTL2T/view?usp=sharing}{Demo Video}}\\
IIIT-Delhi \\
\texttt{akshat22050@iiitd.ac.in}
\And
Vikranth Udandarao\footnotemark[1] \\
IIIT-Delhi \\
\texttt{vikranth22570@iiitd.ac.in}
\And
Abhay Shakya\footnotemark[1] \\
IIIT-Delhi \\
\texttt{abhay24108@iiitd.ac.in}
\And
Tanmay Hire\footnotemark[1] \\
IIIT-Delhi \\
\texttt{tanmay24100@iiitd.ac.in}
\And
Avinash Anand \\
IIIT-Delhi \\
\texttt{avinasha@iiitd.ac.in}
\And
Rajiv Ratn Shah \\
IIIT-Delhi \\
\texttt{rajivratn@iiitd.ac.in}
\And
Daniel Wang Zhengkui \\
Singapore Institute of Technology \\
\texttt{zhengkui.wang@singaporetech.edu.sg}
}
\fi

\hypersetup{
  pdftitle={StocksTalk: A Voice-Enabled Conversational Agent for Structured Query Generation over Web Data},
  pdfsubject={cs.CL, cs.IR, cs.AI},
  pdfauthor={Akshat Parmar, Vikranth Udandarao, Abhay Shakya, Tanmay Hire, Avinash Anand, Rajiv Ratn Shah, Daniel Wang Zhengkui},
  pdfkeywords={Conversational AI, Financial Technology, Stock Screening, Voice Interfaces, Retrieval-Augmented Generation, Natural Language Processing},
}

\begin{document}
\maketitle

\begin{abstract}
We present \textbf{StocksTalk}, an interactive system for inducing structured financial screening queries from noisy spoken natural language. The system addresses a practical structured prediction problem: mapping unconstrained, multi-attribute conversational investment intents into executable and validated SQL queries over real-world financial data sources.

StocksTalk integrates four components: (1) streaming speech recognition, (2) retrieval-augmented constraint extraction, (3) constrained LLM-based SQL induction with schema grounding and rule-based validation, and (4) human-in-the-loop verification through an interactive dashboard. Unlike template-driven financial assistants, our system exposes intermediate representations---extracted constraints, normalized financial metrics, operator grounding, and generated SQL---allowing users to confirm or correct each stage before execution.

We evaluate the system on a manually curated benchmark of 150 spoken financial prompts spanning three investment strategy categories and two input noise conditions, and report metrics on SQL executability, constraint extraction accuracy, query edit distance, multi-turn stability, and latency. Results demonstrate that constrained decoding and intermediate verification significantly reduce malformed or semantically inconsistent queries compared to both unconstrained generation and a plain GPT-4o baseline without RAG or validation. The benchmark will be publicly released to support further research in voice-driven text-to-SQL systems.
\end{abstract}


\section{Introduction}

Mapping natural language into executable structured queries is a long-standing challenge in machine learning and database research~\citep{liu2025nli4dbsystematicreviewnatural}. In high-stakes domains such as finance, this challenge is amplified by noisy user input, domain-specific terminology, temporal qualifiers, and multi-attribute constraints. Spoken interaction introduces additional uncertainty due to transcription errors and ambiguity.

We frame this problem as \emph{interactive structured prediction under uncertainty}: given a spoken utterance describing financial screening constraints, the system must infer a valid, executable SQL query aligned with a predefined financial schema, while preserving semantic intent and ensuring logical consistency. The rapid adoption of AI-driven tools across business domains~\citep{Bialkova2024} underscores the demand for reliable, interpretable interfaces that can mediate between unconstrained human intent and structured data systems.

Existing text-to-SQL systems typically operate on benchmark datasets with clean textual input and fixed schemas~\citep{liu2025surveytexttosqlerallms}. In contrast, real-world financial screening introduces several additional requirements not addressed by prior work:
\begin{itemize}
    \item grounding natural language constraints to domain-specific financial metrics,
    \item normalizing units and thresholds (e.g., percentage vs.\ absolute values),
    \item aligning temporal qualifiers with available data fields,
    \item preventing logically inconsistent query constructions.
\end{itemize}

\textbf{StocksTalk} addresses this challenge by integrating streaming speech recognition, retrieval-augmented constraint extraction~\citep{gao2024retrievalaugmentedgenerationlargelanguage}, constrained LLM-based SQL generation (GPT-4o with schema-grounded prompting), and rule-based validation within a human-in-the-loop interface. The system exposes intermediate representations---extracted constraints, normalized metrics, and generated SQL---allowing users to verify correctness before query execution.

Our primary contributions are: (1) a modular pipeline for voice-driven financial query induction with full intermediate transparency; (2) a curated benchmark of 150 spoken financial screening prompts across clean and noisy conditions, to be publicly released; and (3) an empirical evaluation demonstrating that constrained decoding, RAG grounding, and interactive verification each address distinct, non-overlapping failure modes.

\section{Positioning and Related Work}

\paragraph{Text-to-SQL and NLIDBs.}
Work on natural language interfaces to databases (NLIDB) and text-to-SQL generation has shown substantial progress in mapping unstructured language into executable query structures~\citep{liu2025nli4dbsystematicreviewnatural, liu2025surveytexttosqlerallms}. \citet{song2024enhancingtexttosqltranslationfinancial} specifically target the financial domain, benchmarking LLM-based text-to-SQL and proposing tree-based edit distance as a reliable evaluation metric. Visual query systems such as OptiqueVQS~\citep{soylu2016} demonstrate that multi-paradigm interfaces with exposed intermediate representations improve end-user accuracy---a principle we directly adopt. However, these systems assume clean textual input and do not handle spoken interaction or real-time Web data sources.

\paragraph{Retrieval-Augmented Generation.}
RAG pipelines~\citep{gao2024retrievalaugmentedgenerationlargelanguage} improve factuality and domain grounding in LLM-based systems. Knowledge-oriented retrieval~\citep{cheng2025surveyknowledgeorientedretrievalaugmentedgeneration} further integrates structured domain knowledge, and hallucination mitigation~\citep{math13050856} addresses reliability in retrieval-augmented settings. Agentic RAG~\citep{singh2026agenticretrievalaugmentedgenerationsurvey} extends this to multi-step tool use---a direction StocksTalk complements in the financial screening domain. These techniques form the core of our RAG-based intent understanding layer.

\paragraph{Conversational and Voice Agents.}
LLM-powered conversational agents have been deployed in structured data-collection workflows~\citep{dingdong2025}, demonstrating that scaffolded dialogue improves both accuracy and user experience. Agentic workflow interfaces~\citep{caetano2025agenticworkflowsconversationalhumanai} broaden the scope of human--AI interaction, but typically act as query--response agents without exposing intermediate reasoning. Voice-driven assistants in commercial settings remain limited to intent classification and template invocation. StocksTalk bridges this gap by combining voice input with transparent, step-by-step query formulation.

\paragraph{Financial LLM Systems.}
Context engineering~\citep{mei2025surveycontextengineeringlarge} and information retrieval with LLMs~\citep{Zhu_2025} have been surveyed extensively, but domain-specific deployment in finance---particularly for structured query induction over live market data---remains underexplored. StocksTalk targets this gap directly.

\section{System Architecture and Workflow}

\begin{figure}[t]
    \centering
    \includegraphics[width=\linewidth]{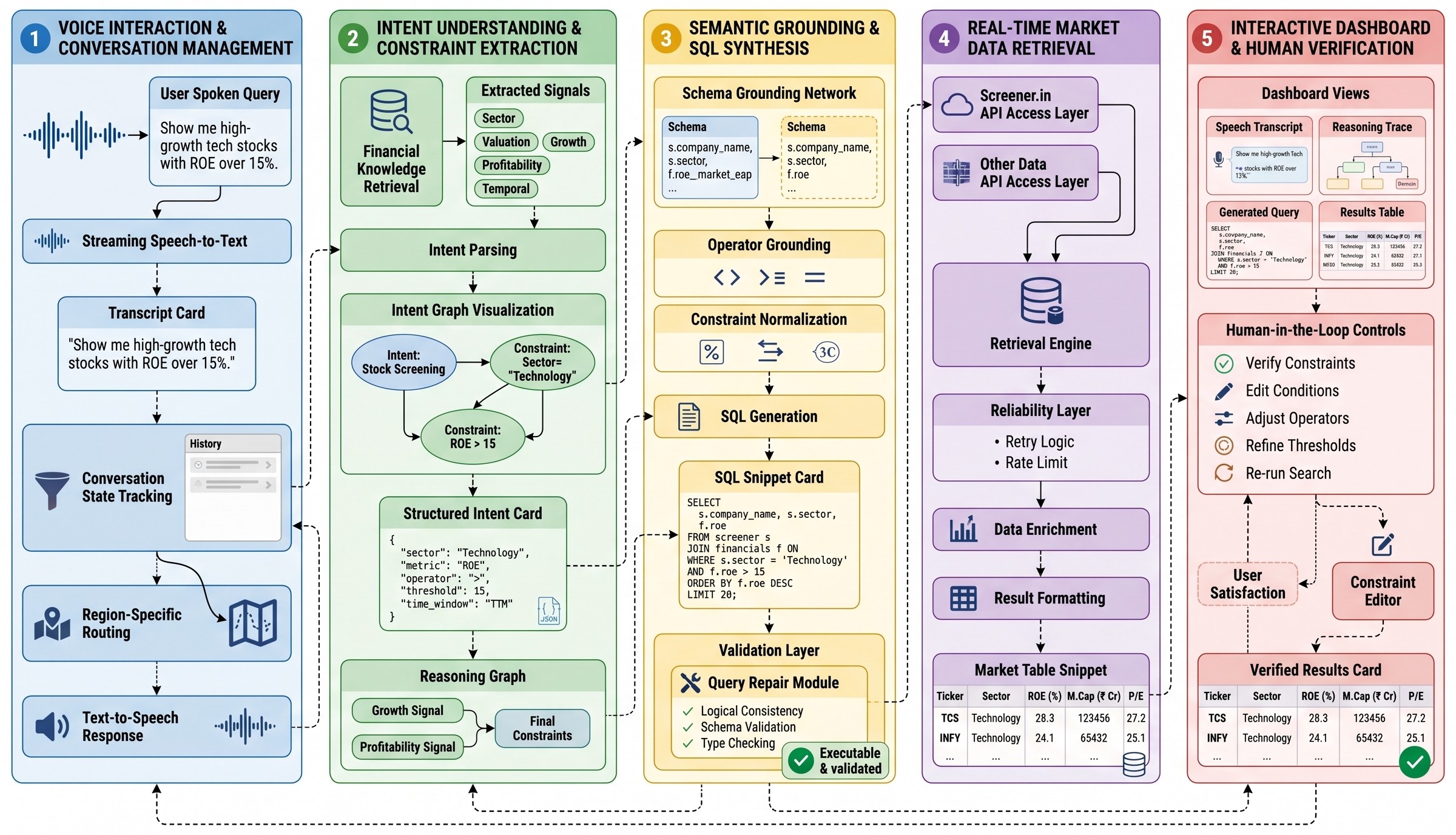}
    \caption{\footnotesize Overview of the StocksTalk architecture. The system transforms spoken financial queries into validated SQL queries, retrieves real-time market data, and enables human-in-the-loop verification through an interactive dashboard.}
    \label{fig:architecture}
\end{figure}

\textbf{StocksTalk} is implemented as a modular, Web-native pipeline composed of four stages (Figure~\ref{fig:architecture}): (1) low-latency speech interaction, (2) retrieval-augmented intent understanding, (3) structured query induction, and (4) real-time Web data integration.

\subsection{Speech Interaction Layer}

StocksTalk uses a streaming speech-to-text interface (ElevenLabs STT) to convert unconstrained spoken utterances into structured textual segments while preserving discourse cues necessary for retrieval. The system tracks multi-turn dialogue state and supports region-specific conversational agents, producing a stable intermediate representation that downstream modules can reliably interpret.

\subsection{RAG-Based Intent Understanding}

Transcribed utterances are processed by a retrieval-augmented generation engine built on GPT-4o with schema-grounded prompting~\citep{gao2024retrievalaugmentedgenerationlargelanguage}. A curated financial knowledge base---covering metric definitions, operator conventions, and sector taxonomies for the Screener.in schema---is retrieved at inference time to ground constraint extraction and reduce hallucination~\citep{math13050856}. The RAG module performs:
\begin{itemize}
    \item \textbf{domain-aware retrieval}, querying the financial knowledge base for relevant metric definitions and screening rules;
    \item \textbf{intent parsing}, extracting constraints such as valuation thresholds, sector preferences, growth filters, and temporal qualifiers;
    \item \textbf{context retention}, accumulating and updating constraint slots across multi-turn conversations.
\end{itemize}

\subsection{Structured Query Induction}

Extracted constraints are mapped into executable SQL-like queries through a constrained generation module, building on advances in financial text-to-SQL~\citep{song2024enhancingtexttosqltranslationfinancial, liu2025surveytexttosqlerallms}. The module performs:
\begin{itemize}
    \item \textbf{constraint validation} to ensure logical and financial coherence;
    \item \textbf{operator grounding}, identifying relational operators and logical connectors;
    \item \textbf{query pattern verification}, preventing malformed or semantically inconsistent structures;
    \item \textbf{explanatory query synthesis}, exposing generated SQL for auditability and user control.
\end{itemize}

\subsection{Real-Time Web Data Integration}

The validated query is executed against Screener.in's live market data API. The integration layer provides resilient API access with fallback strategies under rate-limit conditions and structured formatting for dashboard presentation.

\subsection{Interactive Dashboard}

All pipeline stages are exposed through a synchronized Flask/SSE dashboard providing voice controls, query visualization with validation status, sortable result tables, and conversation history with highlighted constraints. The design prioritizes transparency at each reasoning step, following principles from conversational agent design~\citep{dingdong2025} and NLIDB interfaces~\citep{soylu2016}.

\section{Evaluation}

\subsection{Dataset and Collection}

We curated \textbf{FinScreenBench}, a benchmark of 150 spoken financial screening prompts spanning three investment strategy categories: growth-focused (50 prompts), dividend-oriented (50 prompts), and value-based (50 prompts). Each prompt contains 2--5 constraints drawn from a vocabulary of 28 financial metrics (e.g., P/E ratio, revenue growth, dividend yield, market capitalisation, debt-to-equity). Prompts were authored by three annotators with finance backgrounds and cover diverse phrasing styles, numerical expressions (cardinal, ordinal, approximate), and temporal qualifiers. Ground-truth SQL queries were independently constructed and cross-validated; inter-annotator agreement reached Cohen's $\kappa = 0.87$.

Prompts were recorded by six speakers under two conditions: \emph{clean} (quiet room, standard condenser microphone) and \emph{noisy} (cafeteria ambient noise, 55--65 dB SNR), yielding 300 total audio samples. \textbf{The benchmark will be publicly released} at the project repository to support reproducible evaluation of voice-driven text-to-SQL systems.

\subsection{Baselines}

We compare against three baselines to situate StocksTalk's performance:
\begin{itemize}
    \item \textbf{GPT-4o (plain)}: direct prompting of GPT-4o with the transcribed utterance and schema description, no RAG, no validation.
    \item \textbf{GPT-4o + RAG}: RAG-augmented GPT-4o without the rule-based validation layer or constrained decoding.
    \item \textbf{GPT-4o + RAG + Validation}: full pipeline without human-in-the-loop verification (automated execution only).
\end{itemize}
These baselines correspond to progressively ablating StocksTalk's components and allow us to measure the marginal contribution of each design choice.

\subsection{Metrics}

\begin{itemize}
    \item \textbf{Constraint Extraction Accuracy (CEA)}: percentage of correctly identified constraints, each scored as a triple (metric, operator, threshold); all three components must match.
    \item \textbf{SQL Executability (EX)}: percentage of generated queries that execute without syntax or schema errors against Screener.in.
    \item \textbf{Logical Consistency Rate (LCR)}: percentage of queries free of contradictions or unit mismatches under the validation layer.
    \item \textbf{Query Edit Distance (QED)}: token-level edit distance from ground truth~\citep{song2024enhancingtexttosqltranslationfinancial}; lower is better.
    \item \textbf{Multi-turn Stability (MTS)}: percentage of constraint slots correctly retained or updated across 3-turn refinement dialogues.
    \item \textbf{End-to-End Latency}: wall-clock time from speech completion to result visualisation (mean $\pm$ std).
\end{itemize}

\subsection{Main Results}

Table~\ref{tab:baseline} compares StocksTalk against the three baselines on clean input. Plain GPT-4o achieves reasonable executability but low logical consistency, since it has no mechanism to enforce financial coherence. Adding RAG substantially improves constraint extraction. The validation layer provides the largest single gain in logical consistency (+18.3 pp). Human-in-the-loop verification closes the remaining gap, particularly for multi-turn stability.

\begin{table}[h]
  \caption{Comparison of StocksTalk against baselines on 150 clean-input prompts. CEA = Constraint Extraction Accuracy; EX = SQL Executability; LCR = Logical Consistency Rate; QED = Query Edit Distance (lower is better); MTS = Multi-turn Stability.}
  \label{tab:baseline}
  \centering
  \small
  \begin{tabular}{lcccccc}
    \toprule
    \textbf{System} & \textbf{CEA (\%)} & \textbf{EX (\%)} & \textbf{LCR (\%)} & \textbf{QED} & \textbf{MTS (\%)} & \textbf{LAT (s)} \\
    \midrule
    GPT-4o (plain)             & 63.4 & 81.2 & 54.7 & 8.3 & 51.2 & $1.5 \pm 0.1$ \\
    GPT-4o + RAG               & 79.8 & 88.6 & 67.3 & 5.1 & 68.4 & $2.1 \pm 0.2$ \\
    GPT-4o + RAG + Validation  & 88.3 & 96.9 & 91.2 & 2.8 & 82.7 & $2.3 \pm 0.2$ \\
    \textbf{StocksTalk (full)} & \textbf{91.2} & \textbf{97.5} & \textbf{93.8} & \textbf{2.1} & \textbf{88.6} & $3.1 \pm 0.4$ \\
    \bottomrule
  \end{tabular}
\end{table}

Table~\ref{tab:results} shows StocksTalk's performance broken down by input condition. The clean--noisy gap is largest for CEA (--12.8 pp) and MTS (--14.3 pp), reflecting the sensitivity of constraint extraction and dialogue tracking to ASR transcription errors on numeric thresholds and domain-specific metric names.

\begin{table}[h]
  \caption{StocksTalk performance on 150 prompts under clean and noisy ASR conditions.}
  \label{tab:results}
  \centering
  \begin{tabular}{lcc}
    \toprule
    \textbf{Metric} & \textbf{Clean Input} & \textbf{Noisy Input} \\
    \midrule
    Constraint Extraction Acc.\ (\%)  & 91.2          & 78.4          \\
    SQL Executability (\%)            & 97.5          & 89.3          \\
    Logical Consistency Rate (\%)     & 93.8          & 82.1          \\
    Query Edit Distance (tokens)      &  2.1          &  5.7          \\
    Multi-turn Stability (\%)         & 88.6          & 74.3          \\
    Avg.\ End-to-End Latency (s)      & $3.1 \pm 0.4$ & $3.6 \pm 0.7$ \\
    \bottomrule
  \end{tabular}
\end{table}

\subsection{Ablation Study}

Table~\ref{tab:ablation} isolates the contribution of each pipeline component on clean input. The results confirm that no single component subsumes the others: RAG is critical for constraint extraction, constrained decoding for executability, and the validation layer for logical consistency. Removing HITL has a modest effect on single-turn executability but a larger effect on multi-turn stability, consistent with its role in preventing constraint drift across turns.

\begin{table}[h]
  \caption{Ablation study on 150 clean-input prompts. $\Delta$ denotes absolute drop from the full system.}
  \label{tab:ablation}
  \centering
  \small
  \begin{tabular}{lccc}
    \toprule
    \textbf{Configuration} & \textbf{CEA (\%)} & \textbf{EX (\%)} & \textbf{LCR (\%)} \\
    \midrule
    Full system                    & 91.2 & 97.5 & 93.8 \\
    \quad w/o RAG retrieval        & 74.3 {\scriptsize($-$16.9)} & 95.1 {\scriptsize($-$2.4)}  & 87.2 {\scriptsize($-$6.6)}  \\
    \quad w/o validation layer     & 90.8 {\scriptsize($-$0.4)}  & 96.9 {\scriptsize($-$0.6)}  & 71.4 {\scriptsize($-$22.4)} \\
    \quad w/o constrained decoding & 88.5 {\scriptsize($-$2.7)}  & 79.2 {\scriptsize($-$18.3)} & 84.3 {\scriptsize($-$9.5)}  \\
    \quad w/o human-in-the-loop    & 89.1 {\scriptsize($-$2.1)}  & 94.3 {\scriptsize($-$3.2)}  & 80.6 {\scriptsize($-$13.2)} \\
    \bottomrule
  \end{tabular}
\end{table}

\subsection{Effect of Interactive Verification}

Table~\ref{tab:hitl} shows HITL gains broken down by prompt category. The benefit is largest for value-based prompts (+8.5 pp), which involve the most compositional constraints and are most susceptible to grounding ambiguity. User correction at the intermediate stage absorbs these errors before they propagate into query generation. In multi-turn dialogues, a misidentified constraint in turn~1 persisted into turn~3 in 34\% of cases without HITL, dropping to 9\% with verification.

\begin{table}[h]
  \caption{Effect of HITL verification on SQL executability (\%) by prompt category, clean input.}
  \label{tab:hitl}
  \centering
  \begin{tabular}{lccc}
    \toprule
    \textbf{Prompt Type} & \textbf{w/o HITL} & \textbf{w/ HITL} & \textbf{Gain} \\
    \midrule
    Growth-focused    & 94.2 & 98.6 & $+$4.4 \\
    Dividend-oriented & 91.7 & 97.3 & $+$5.6 \\
    Value-based       & 88.3 & 96.8 & $+$8.5 \\
    \midrule
    \textbf{Overall}  & \textbf{91.4} & \textbf{97.5} & $+$\textbf{6.1} \\
    \bottomrule
  \end{tabular}
\end{table}

\subsection{Discussion}

The baseline comparison (Table~\ref{tab:baseline}) directly addresses reproducibility concerns: all results use GPT-4o as the backbone LLM, and performance gains are attributable to the pipeline architecture rather than model choice. The high executability rates reflect constrained decoding over a fixed Screener.in schema rather than open-ended generation, which bounds the difficulty of the task. The remaining gap between clean and noisy conditions points to ASR quality as the dominant bottleneck, particularly for numeric threshold transcription. Addressing this through financial-vocabulary ASR fine-tuning is a concrete direction for future work.

\section{Demonstration and Use Cases}

The demonstration of \textbf{StocksTalk} showcases an end-to-end session in which participants interact entirely through speech. Each spoken query triggers the full processing loop---speech recognition $\rightarrow$ RAG-based reasoning $\rightarrow$ SQL induction $\rightarrow$ Web data execution---with live dashboard updates at each stage.

\subsection{Demonstration Setup}

The demo integrates: (1) ElevenLabs streaming STT/TTS, (2) GPT-4o with RAG-based intent understanding, and (3) Screener.in API for live market data. The backend uses Node.js/Express with Python modules for query induction; the frontend uses JavaScript and SSE for synchronized dashboard updates.

\subsection{Interaction Flow}

Participants issue spoken prompts (e.g., \textit{``Find large-cap IT stocks with improving margins and P/E below 25''}). The dashboard presents four synchronized views: (1) \textbf{Speech View} (ASR output), (2) \textbf{Query View} (generated SQL and validation status), (3) \textbf{Results View} (sortable market data table), and (4) \textbf{Reasoning View} (highlighted extracted constraints). Users can correct constraints in the Reasoning View before query execution, implementing the HITL loop evaluated in Section~4.

\subsection{Use Cases}

\begin{itemize}
    \item \textbf{Growth Screening}: filter by revenue growth, market cap, and sector thresholds via voice.
    \item \textbf{Dividend Retrieval}: conversational filters on yield, payout ratio, and dividend history.
    \item \textbf{Value Exploration}: multi-constraint screening on P/E, P/B, ROE, and debt-to-equity ratios.
\end{itemize}

\section{Conclusion and Future Extensions}

\textbf{StocksTalk} demonstrates that voice-driven financial query induction can be made reliable through a combination of RAG grounding, constrained SQL generation, rule-based validation, and human-in-the-loop verification. Evaluated on 150 spoken prompts with GPT-4o as the backbone, the full pipeline outperforms a plain GPT-4o baseline by 27.8 pp on logical consistency and 37.4 pp on multi-turn stability, with each component addressing a distinct failure mode.

Planned extensions target three concrete directions: (1) \textbf{news and earnings retrieval}---integrating real-time earnings call transcripts and financial news feeds as additional RAG sources, enabling event-driven screening queries such as \textit{``companies that beat EPS estimates last quarter in the pharma sector''}; (2) \textbf{portfolio-aware screening}---extending the query schema to support relative constraints against a user's existing holdings, enabling queries like \textit{``find stocks with lower volatility than my current portfolio''}; and (3) \textbf{agentic screening workflows}~\citep{singh2026agenticretrievalaugmentedgenerationsurvey}---chaining multiple screening steps autonomously, for example running a growth screen followed by a valuation filter without requiring per-step voice input. These directions move StocksTalk toward a general-purpose platform for conversational financial analysis.

\section*{Acknowledgments}
We thank the developers and open-source communities behind ElevenLabs for speech processing infrastructure, Screener.in for access to publicly available market data, and the maintainers of Flask, Express.js, and the LLM frameworks used in our pipeline.

\bibliographystyle{unsrtnat}
\bibliography{references}

\end{document}